\documentclass{article}

\usepackage{PRIMEarxiv}

\usepackage[utf8]{inputenc} % allow utf-8 input
\usepackage[T1]{fontenc}    % use 8-bit T1 fonts
\usepackage{hyperref}       % hyperlinks
\usepackage{url}            % simple URL typesetting
\usepackage{booktabs}       % professional-quality tables
\usepackage{amsfonts}       % blackboard math symbols
\usepackage{nicefrac}       % compact symbols for 1/2, etc.
\usepackage{microtype}      % microtypography
\usepackage{lipsum}
\usepackage{fancyhdr}       % header
\usepackage{graphicx}       % graphics
\usepackage{float}
\graphicspath{{media/}}     % organize your images and other figures under media/ folder

\title{Correlation Vs Causation in Alzheimer’s Disease: An Interpretability-Driven Study
}

\author{
  Hamzah Dabool \\
  Big Data Analytics Center, College of Information Technology \\
  United Arab Emirates University \\
  Alain, UAE\\
  \texttt{
hamzah.dabool@uaeu.ac.ae} \\
   \And
  Raghad Mustafa \\
  Faculty of Information Technology Engineering \\
  Damascus University\\
 Damascus, Syria\\
  \texttt{raghad6475@gmail.com} \\
}

\begin{document}
\maketitle

\begin{abstract}
Understanding the distinction between causation and correlation is critical in Alzheimer’s disease (AD) research, as it impacts diagnosis, treatment, and the identification of true disease drivers. This experiment investigates the relationships among clinical, cognitive, genetic, and biomarker features using a combination of correlation analysis, machine learning classification, and model interpretability techniques. Employing the XGBoost algorithm, we identified key features influencing AD classification, including cognitive scores and genetic risk factors. Correlation matrices revealed clusters of interrelated variables, while SHAP (SHapley Additive exPlanations) values provided detailed insights into feature contributions across disease stages. Our results highlight that strong correlations do not necessarily imply causation, emphasizing the need for careful interpretation of associative data. By integrating feature importance and interpretability with classical statistical analysis, this work lays groundwork for future causal inference studies aimed at uncovering true pathological mechanisms. Ultimately, distinguishing causal factors from correlated markers can lead to improved early diagnosis and targeted interventions for Alzheimer’s disease.
\end{abstract}

% keywords can be removed

\section{Introduction}
Alzheimer’s disease (AD) is a progressive neurological disorder that causes brain cells to degenerate and die, leading to a continuous decline in memory, thinking, behavior, and the ability to perform everyday tasks \cite{ballard2011alzheimer}. It is the most common cause of dementia among older adults \cite{hendrie1998epidemiology}. The disease typically begins with mild memory loss and confusion, often mistaken for normal aging. It progresses through several stages: early-stage (mild), where individuals may still function independently but have noticeable memory lapses; middle-stage (moderate), where symptoms become more pronounced, including difficulty recognizing people, disorientation, and changes in personality or behavior; and late-stage (severe), where individuals lose the ability to respond to their environment, carry on a conversation, or control movement \cite{caselli2012characterizing}. But As the disease advances, full-time care is usually required.

Multi-class classification is a supervised learning task where a model is trained to categorize input data into one of three or more distinct classes \cite{rifkin2008multiclass}. Unlike binary classification, which deals with only two categories, multi-class classification handles problems where each instance belongs to one class out of several possible outcomes. This approach is commonly applied in AD classification to assign patients to multiple diagnostic categories—such as cognitively normal, mild cognitive impairment, and Alzheimer’s disease—based on various clinical, cognitive, and biomarker features\cite{farooq2017deep}.

Causation refers to the identification of cause-and-effect relationships between variables, where changes in one feature directly influence changes in another. Understanding causation allows models to predict the outcomes of interventions or changes, rather than simply recognizing patterns in data. Establishing causation typically requires careful experimental design, causal inference methods, or domain knowledge to move beyond associations captured by standard predictive models \cite{marwala2015causality}.

Correlation in machine learning describes a statistical relationship where two or more variables change together, either positively or negatively, without necessarily implying a direct cause-effect link. Correlations help models detect patterns and dependencies in data, which improve prediction accuracy. However, correlation alone does not indicate whether one variable causes another, and relying solely on correlated features can lead to misleading conclusions if causal relationships are not considered \cite{marwala2015causality}.

This paper contributes to Alzheimer’s disease research by systematically analyzing the relationships between key clinical, cognitive, and genetic features through both correlation and causation perspectives. Using machine learning techniques, including XGBoost and SHAP interpretability, we identify the most influential features for multi-class disease classification while highlighting the limitations of correlation-based insights. By integrating statistical correlation analysis with model-driven feature importance, this work lays a foundation for future causal inference studies aimed at uncovering true disease drivers, ultimately supporting improved diagnosis and targeted interventions in Alzheimer’s disease.

The rest of the paper is organized as follows: Section 2 outlines the methodology, including the motivation, dataset description, and machine learning settings. Section 3 details the data preprocessing steps and model training, along with feature importance analysis, correlation calculations, and SHAP value computations. In Section 4, we discuss the results, and finally, Section 5 concludes the paper.

\section{Methodology}
This section outlines the motivation driving the experimental study and provides an overview of the selected model along with its configuration, prior to proceeding with the implementation. 
\subsection{Motivation}
The motivation behind this implementation is to explore the relationship between causation and correlation in the context of dementia. The aim is to investigate whether a meaningful connection exists between the two or if they operate independently. The key findings from this experiment are expected to provide valuable insights that will inspire and guide future scientific research in this area.

\subsection{Dataset}
In this experiment, we utilized a comprehensive dataset formed by combining three key sources: ADNI1, ADNI2, and ADNIGo. This unified dataset supports a detailed exploration across five diagnostic categories: Cognitively Normal (CN), Alzheimer’s Disease (AD), Late Mild Cognitive Impairment (LMCI), Significant Memory Concern (SMC), and Early Mild Cognitive Impairment (EMCI). It contains 13,900 records and 41 features, encompassing both numerical and categorical variables. Despite its richness, the dataset presents a major hurdle—approximately 250,480 values are missing, accounting for about 43\% of the entire data, which introduces complexity in data processing and analysis.

\subsection{Machine Learning Model}\label{sss}
The chosen method for this experiment is the Extreme Gradient Boosting (XGBoost) algorithm, as it has demonstrated strong classification performance on this dataset \cite{dabool2024enhancing}. Additionally, XGBoost is well-suited for medical data because it can handle missing values internally, eliminating the need for imputation techniques that might compromise the integrity of sensitive medical information \cite{dabool2024enhancing}.

The hyperparameter values of the XGBoost model will be set  as ’gamma’ = 0.1, ’eta’ = 0.1, ’max depth’ = 0, and ’min child weight’ = 0 as these values were proven to increase the performance of the model after tuned using GridSearch tuning with 5 fold cross validation scoring 75.96\% validation accuracy score \cite{dabool2024comparative}.

\section{Implementation and results}

\subsection{Data preparation}
At the start of our study, the dataset contained 13,900 records. Among them, 20 entries were missing class labels and were therefore removed, reducing the total to 13,880 samples. Additionally, 2,277 duplicate records were identified and eliminated, resulting in a final working dataset of 11,603 unique samples.

\subsection{XGboost Model Initiation}\label{ddd}
The model, as described in Section \ref{sss}, was trained and evaluated using 5-fold cross-validation. It achieved an accuracy of 75.96\%, aligning with our expectations and confirming its readiness for the next stages of the experiment.

\subsection{Features Importance}
After initiating the model in Section~\ref{ddd}, we analyzed the top 10 features with the highest importance. These features play a critical role in the internal decision-making of the XGBoost model, strongly influencing how the decision trees split and ultimately determining the final classification outcome.

\begin{figure}[H]
\centering
\includegraphics[width=140mm]{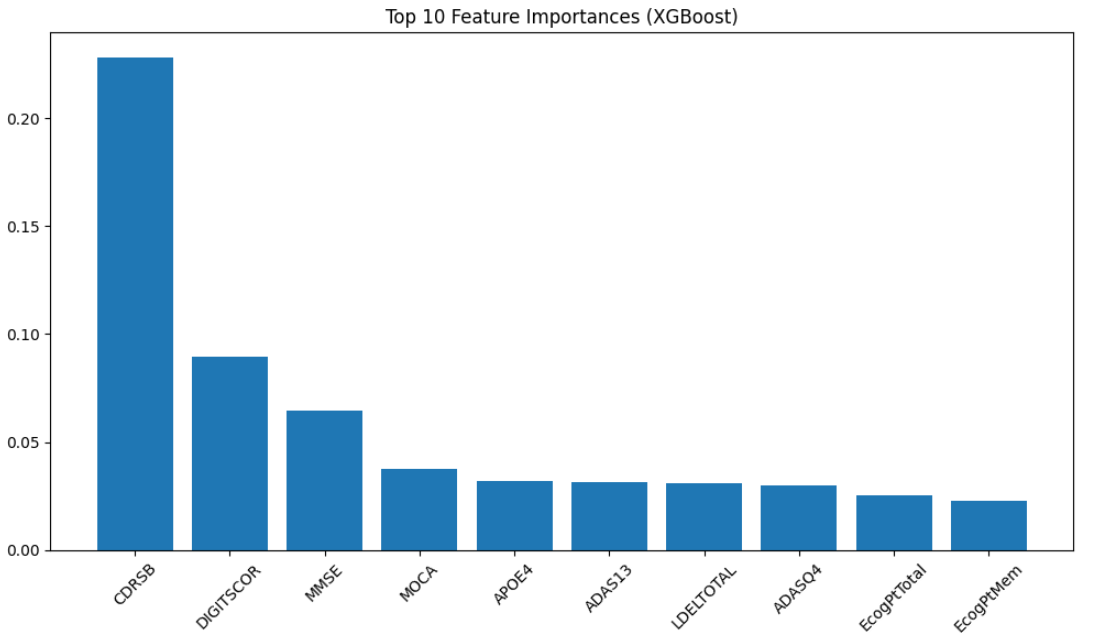}
\caption{Feature importances derived from the XGBoost model.}
\label{fig:raghad_importances}
\end{figure}

The bar chart in Figure \ref{fig:raghad_importances} illustrates the most influential features identified by the XGBoost model during the classification process. The horizontal axis lists the top 10 features, while the vertical axis shows their corresponding importance scores, which reflect how much each feature contributes to the model's decision-making. Among these, CDRSB (Clinical Dementia Rating Sum of Boxes) stands out with the highest importance, indicating it plays a dominant role in predicting the outcome. It is followed by DIGITSCORE, MMSE (Mini-Mental State Examination), and MOCA (Montreal Cognitive Assessment), which are all cognitive assessment tools commonly used in dementia evaluation. Other features like APOE4 (a genetic risk factor), ADAS13, and various scores related to memory and executive function also contribute meaningfully. This ranking provides insight into which clinical and cognitive variables are most predictive in the context of AD classification.

\subsection{Correlation Matrix}
We calculated the correlation coefficients between every pair of features in the dataset and presented these relationships visually through a detailed heatmap, enabling easy identification of strong positive or negative associations among variables.

\begin{figure}[H]
\centering
\includegraphics[width=140mm]{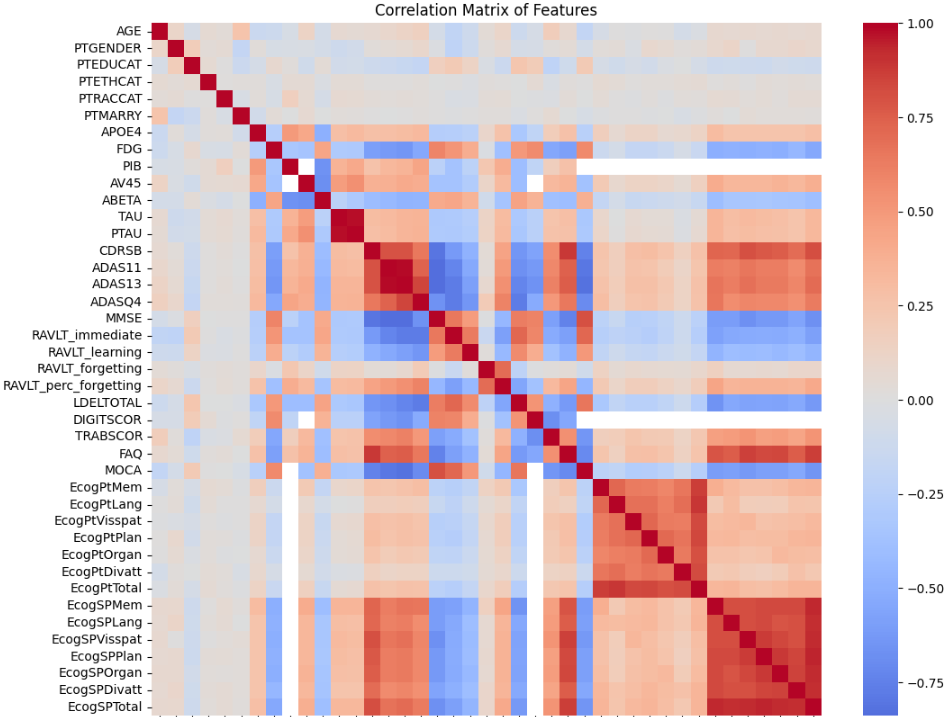}
\caption{Correlation Matrix derived from the XGBoost model with respect to the classes available.}
\label{fig:raghad_correlation}
\end{figure}

The heatmap in Figure \ref{fig:raghad_correlation} displays the pairwise correlation coefficients among various clinical, demographic, genetic, and cognitive features in the dataset. Each cell represents the strength and direction of the linear relationship between two variables, with red tones indicating positive correlations and blue tones indicating negative correlations. Strong correlations appear closer to deep red (value near +1) or deep blue (value near –1), while weak or no correlation appears in white or pale colors (value near 0). For example, a strong positive correlation is observed among several Ecog (Everyday Cognition) scores, suggesting they measure related cognitive abilities. Similarly, biomarkers such as TAU, PTAU, and ABETA show interdependencies. This matrix is useful for identifying redundant features, understanding feature relationships, and guiding feature selection or dimensionality reduction in downstream modeling tasks.

\subsection{SHAP}
Finally, we computed the average absolute SHAP values for each feature, breaking down their impact individually across all classes to better understand how each feature contributes to the model’s predictions within different classification groups.ss

\begin{figure}[H]
\centering
\includegraphics[width=140mm]{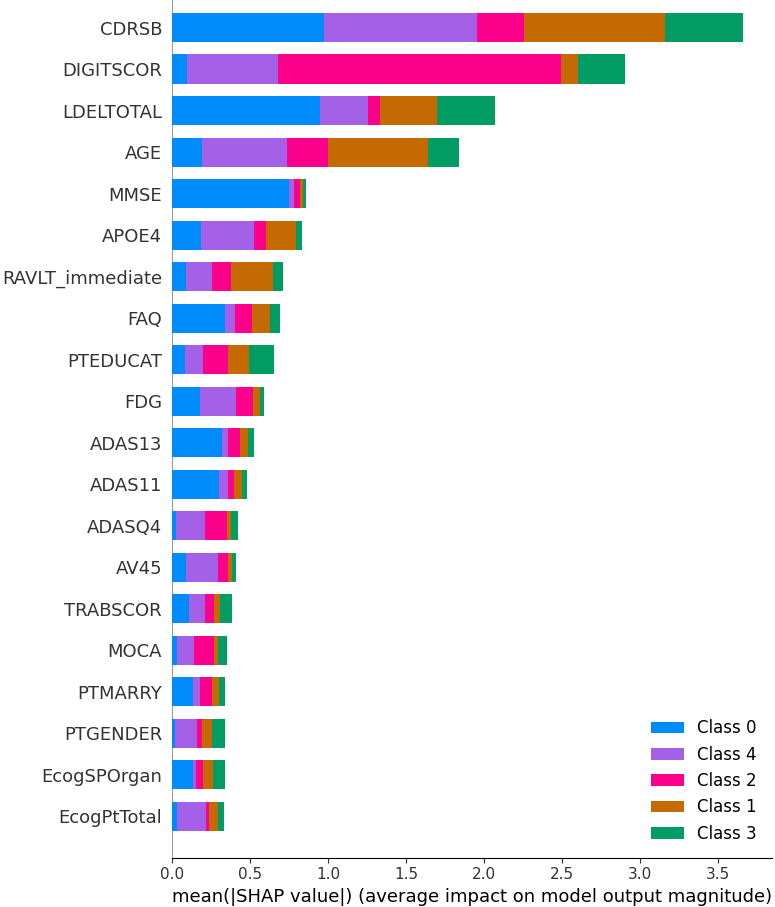}
\caption{Mean SHAP value of all features with respect to the classes available.}
\label{fig:raghad_shap}
\end{figure}

The stacked bar chart in Figure \ref{fig:raghad_shap} displays the average absolute SHAP values for the top features, indicating their overall impact on the model's predictions across different classes. Each feature is represented on the vertical axis, while the horizontal axis quantifies the mean magnitude of its contribution to the model output. The colored segments within each bar correspond to the influence of the feature on specific classes, with the legend identifying five classes labeled 0 through 4. For example, CDRSB shows the highest overall impact, with substantial contributions across multiple classes, particularly Class 0 and Class 1. Other cognitive and demographic features, such as DIGITSCOR, LDELTOTAL, and AGE, also demonstrate significant influence distributed variably among classes. This visualization helps to understand not only which features are most important globally but also how their effects differ depending on the predicted class.

\section{Discussion}
Understanding the complex relationships between clinical, cognitive, and biological features in AD is crucial for improving early diagnosis and developing targeted interventions. In this study, we aimed to investigate these relationships from both correlation and causation perspectives, utilizing machine learning interpretability tools alongside classical statistical analyses.

\subsection{Feature Importance and Predictive Contributions}
Figure 1 displays the top 10 features ranked by their importance in the XGBoost classification model. The Clinical Dementia Rating Sum of Boxes (CDRSB) emerges as the most critical predictor, highlighting its strong association with disease progression. Other prominent features include cognitive scores such as DIGITSCOR, MMSE, and MOCA, which are widely recognized clinical assessments reflecting cognitive decline in AD. Genetic factors like APOE4, a well-established risk allele, and neuropsychological test scores such as ADAS13 also contribute significantly. These results underscore that a combination of clinical cognitive measures and genetic predisposition are pivotal in the model’s ability to classify disease status accurately.

\subsection{Correlation Patterns Among Feature}
The correlation matrix (Figure 2) provides insight into the linear associations between features, revealing clusters of highly correlated variables. For instance, cognitive assessment scores and functional evaluations such as ADAS and Ecog metrics show strong positive correlations, likely due to measuring overlapping domains of cognitive and functional impairment. Biomarkers including TAU, PTAU, and ABETA exhibit complex interrelations consistent with known AD pathophysiology. However, while these correlations indicate relationships, they do not inherently imply causal links. For example, the high correlation between cognitive test scores may reflect shared underlying neuropathology, but each score's individual contribution to disease progression requires further investigation.

\subsection{SHAP Value Decomposition and Class-Specific Impacts}
Figure 3 decomposes the average absolute SHAP values by class, illustrating how each feature influences model predictions across different diagnostic categories. This nuanced view reveals that while features like CDRSB and DIGITSCOR consistently impact all classes, their relative influence varies, reflecting heterogeneity in disease presentation and progression. This class-wise interpretability is critical for understanding the differential roles features may play in early versus late stages of cognitive decline and across distinct patient subgroups.

\subsection{Correlation Versus Causation: Interpretative Implications}
The juxtaposition of correlation and feature importance analyses highlights a fundamental challenge in AD research: distinguishing mere associations from causal mechanisms. Correlation analyses, while useful for identifying related variables and potential biomarkers, cannot establish cause-effect relationships due to confounding factors and bidirectional influences inherent in complex biological systems. The strong correlations among cognitive and biomarker features underscore the interconnected nature of AD pathology but leave open the question of which factors are drivers versus passengers.

Machine learning models such as XGBoost, complemented by SHAP explanations, offer powerful tools for uncovering feature relevance in prediction tasks but do not, on their own, confirm causality. They effectively identify features that influence model output, potentially flagging variables with causal roles or strong proxies. Yet, high feature importance or SHAP values reflect predictive power rather than mechanistic influence.

To move from correlation to causation, integrating these findings with domain knowledge and applying causal inference frameworks is essential. For example, controlled longitudinal studies, Mendelian randomization, or causal graphical models could help disentangle direct effects from confounding. Our results provide a data-driven foundation identifying key features to focus on in such causal analyses.

\subsection{Exploring the Link Between Correlation and Causation in Alzheimer’s Disease Features}

Our analyses reveal a complex interplay between correlated features and their potential causal influence on Alzheimer’s disease classification. The correlation matrix (Figure 2) shows several strong associations among cognitive assessments, clinical scores, and biomarker levels, indicating that many features move together as the disease progresses. For instance, multiple cognitive tests and functional evaluations cluster tightly, reflecting overlapping constructs of cognitive decline.

However, these correlations alone do not clarify whether these features are causal agents driving disease pathology or simply co-varying markers reflecting the same underlying processes. The feature importance ranking (Figure 1) and SHAP value breakdown by class (Figure 3) help distinguish which features most influence the model’s predictions, identifying variables like CDRSB and DIGITSCOR as key drivers in classification. These results suggest these features capture critical patterns relevant to distinguishing disease stages.

Interestingly, while some highly correlated features such as MMSE and MOCA contribute meaningfully to predictions, their predictive power may arise from reflecting downstream consequences of disease progression rather than initiating causes. The variability in feature impact across classes further underscores this complexity—some features exert stronger influence in specific disease states, hinting at stage-dependent roles.

This divergence between correlation strength and predictive importance underscores a crucial point: features strongly correlated with diagnosis or each other may not necessarily be the primary causal factors but could act as proxies or consequences. Conversely, features with moderate correlations but high importance may capture subtler yet critical signals linked to causal mechanisms.

\section{Conclusion}
This study highlights the complex relationship between correlation and causation in understanding Alzheimer’s disease. Through comprehensive analyses combining correlation matrices, machine learning feature importance, and SHAP-based interpretability, we identified key clinical, cognitive, and genetic features strongly associated with disease classification. While correlation analyses revealed groups of related variables, feature importance metrics pointed out those most influential in predictive modeling. Crucially, our findings emphasize that strong correlation does not mean causation; many features may serve as markers or results of underlying pathology rather than direct causes.Separating true causes from correlated factors remains essential for advancing Alzheimer’s research and improving clinical care. Our work provides a foundation for future studies using careful causal inference methods to confirm and build on these insights. By combining data-driven approaches with expert knowledge and causal frameworks, we can better identify real disease mechanisms, improve early diagnosis, and develop more targeted treatments, ultimately enhancing patient outcomes.

\bibliographystyle{unsrt}  
% \bibliography{references}  

\end{document}